\documentclass[11pt,a4paper]{article}
\usepackage[hyperref]{acl2020}
\usepackage{times}
\usepackage{latexsym}
\usepackage{bm}
\usepackage{stfloats}
\usepackage{subcaption}

\usepackage[T5,T1]{fontenc}

\DeclareTextSymbolDefault{\ohorn}{T5}
\DeclareTextSymbolDefault{\uhorn}{T5}

\usepackage{tikz-dependency}

\newcommand{\BERT}{\mathrm{BERT}}

\usepackage{microtype}
\usepackage{amsmath}
\usepackage{xspace}
\usepackage{amsfonts}
\usepackage{amsthm}
\usepackage{amssymb}
\usepackage{comment}
\usepackage{adjustbox}
\usepackage{tipa}

\newcommand{\word}[1]{\textit{#1}}

\newtheorem{prop}{Proposition}

\usepackage{booktabs}
\usepackage{cleveref}
\crefname{section}{\S}{\S\S}
\Crefname{section}{\S}{\S\S}
\crefname{table}{Table}{}
\crefname{figure}{Figure}{}
\crefname{algorithm}{Algorithm}{}
\crefname{equation}{eq.}{}
\crefname{appendix}{App.}{}
\crefname{prop}{Prop.}{}
\crefformat{section}{\S#2#1#3}

\everypar{\looseness=-1}

\usepackage[disable]{todonotes}

\makeatother

\aclfinalcopy 

\newcommand{\defn}[1]{\textbf{#1}}
\newcommand{\vw}{\mathbf{w}}
\newcommand{\dist}{\Delta}
\newcommand{\dB}{d_B}
\newcommand{\vt}{\mathbf{t}}
\newcommand{\vh}{\mathbf{h}}
\newcommand{\calT}{\mathcal{T}}

\newcommand{\R}{\mathbb{R}}

\newcommand{\calD}{\mathcal{D}}
\newcommand{\vx}{\mathbf{x}}
\newcommand{\Lhm}{\mathcal{L}}
\newcommand{\lhm}{\ell}

\newcommand{\Lperc}{\mathcal{L}}
\newcommand{\lperc}{\ell}

\newcommand{\saveForCR}[1]{}

\title{A Tale of a Probe and a Parser}

\newcommand{\ucambridge}{\normalfont \text{\textipa{D}}}
\newcommand{\ethz}{\text{\normalfont \textipa{Q}}}
\newcommand{\fairesearch}{\normalfont \text{\textipa{@}}}

\author{Rowan Hall Maudslay$^{\ucambridge}$ Josef Valvoda$^{\ucambridge}$ Tiago Pimentel$^{\ucambridge}$ \\
\textbf{Adina Williams$^{\fairesearch}$ Ryan Cotterell$^{\ucambridge,\ethz}$} \\
  $^{\ucambridge}$University of Cambridge~\;~$^{\fairesearch}$Facebook AI Research~\;~%
  $^{\ethz}$ETH Z\"{u}rich \\
  \texttt{rh635@cam.ac.uk},~\;~ \texttt{jv406@cam.ac.uk},~\;~ \texttt{tp472@cam.ac.uk}, \\
  \texttt{adinawilliams@fb.com},~\;~ \texttt{ryan.cotterell@inf.ethz.ch}
}

\pgfkeys{%
/depgraph/reserved/edge style/.style = {%
-, >=stealth, 
},%
}

\begin{document}
\maketitle
\begin{abstract}
Measuring what linguistic information is encoded in 
neural models 
of language 
has become 
popular 
in NLP. 
Researchers approach this enterprise by training ``probes''---supervised models designed to 
extract linguistic structure from another model's output. 
One such probe is the structural probe \cite{hewitt}, designed to quantify the extent to which syntactic information is encoded in contextualised word representations. 
The structural probe has a novel design, unattested in the parsing literature, the precise benefit of which is not immediately obvious. 
To explore whether syntactic probes would do better to 
make use of existing techniques, 
we compare the structural probe to a more traditional parser with an \emph{identical} lightweight parameterisation. The parser outperforms structural probe on UUAS in seven of nine analysed languages, often by a substantial amount (e.g.\ by $11.1$ points in English). Under a second less common metric, however, 
there is the opposite trend---the structural probe outperforms the parser. This begs the question: which metric should we prefer?
\looseness=-1
\end{abstract}

\section{Introduction}
Recently, unsupervised sentence encoders such as ELMo \citep{peters-etal-2018-deep} and $\BERT{}$ \citep{devlin-etal-2019-bert} have become popular within NLP. 
These pre-trained models boast impressive 
performance 
when used in 
many language-related tasks, but 
this gain 
has come at the cost of interpretability. A natural question to ask, then, is whether these models encode the traditional linguistic structures one might expect, such as part-of-speech tags or dependency trees. 
To this end, researchers have invested in the design of diagnostic tools commonly referred to as \defn{probes} \cite{alain2016understanding, conneau-etal-2018-cram, hupkes2018visualisation, poliak-etal-2018-collecting,  marvin-linzen-2018-targeted, niven-kao-2019-probing}. Probes are supervised models designed to extract a target linguistic structure from the output representation learned by another model. \looseness=-1

Based on the authors' reading of the probing literature, there 
is little consensus on where to draw the line between probes and models for performing a target task (e.g.\ a part-of-speech tagger versus a probe for identifying parts of speech). The main distinction 
appears to be 
one of researcher intent: probes are, in essence, a visualisation method \citep{hupkes2018visualisation}. Their goal is not to best the state of the art, but rather to indicate whether certain information is readily available in a model---probes should not ``dig'' for information, they should just expose what is already present. Indeed, a sufficiently expressive probe with enough training data could learn \emph{any} task \cite{hewitt-liang-2019-designing}, but 
this 
tells us nothing about a representation, so it is beside the point. 
For this reason, probes are made ``simple'' \citep{liu-etal-2019-linguistic}, which 
usually means they are minimally parameterised.\footnote{An information-theoretic take on probe complexity is the subject of concurrent work; see \newcite{pimentel-etal-2020-information}.}

Syntactic probes, then, are designed to measure the extent to which a target 
model encodes syntax. A popular example is the \defn{structural probe} \citep{hewitt}, used to compare the syntax that is decodable from different contextualised word embeddings. 
Rather than adopting methodology from the parsing literature, this probe utilises a novel approach for syntax extraction. However, the precise motivation for this novel approach is not immediately clear, since it has nothing to do with model complexity, and appears orthogonal to the goal of a probe. 
Probes are designed to help researchers understand what information exists in a model, and unfamiliar ways of measuring this information may obscure whether we are actually gaining an insight about the representation we wish to examine, or the tool of measurement itself.\looseness=-1

Using the structural probe as a case study, we explore whether 
there is merit in designing models specifically for the purpose of 
probing---whether we should distinguish between the fundamental \emph{design} of probes and models for performing an equivalent task, as opposed to 
just comparing their simplicity. 
We pit the structural probe against a simple parser that has the 
exact same lightweight parameterisation, but instead employs a standard loss function for parsing.  Experimenting on multiligual $\BERT{}$ \citep{devlin-etal-2019-bert}, we find that in seven of nine typologically diverse languages studied (Arabic, Basque, Czech, English, Finnish, Japanese, Korean, Tamil, and Turkish), the parser boosts UUAS dramatically; for example, we observe an $11.1$-point improvement in English.

In addition to using UUAS, \newcite{hewitt} also introduce 
a new metric---correlation of pairwise distance predictions with the gold standard. 
We find that the structural probe outperforms the more traditional parser substantially in terms of this new metric, but it is unclear why this metric matters more than UUAS. 
In our discussion, we contend that, unless a convincing argument to the contrary is provided, traditional metrics  
are preferable. 
Justifying metric choice is of central importance for probing, lest we muddy the waters with a preponderance of ill-understood metrics.

\section{Syntactic Probing Using Distance} \label{sec:synprobing}
Here we introduce \defn{syntactic distance}, which we will later train a probe to approximate.

 \begin{figure}
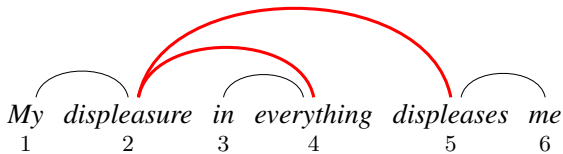

\resizebox{\columnwidth}{!}{%
\begin{dependency}[theme=simple] 
\begin{deptext}[column sep=0.1cm]

		\word{My} \& \word{displeasure} \& \word{in} \& \word{everything} \& \word{displeases} \& \word{me} \\
		{\small $1$} \& 	{\small $2$} \& 	{\small $3$} \& 	{\small $4$} \& 	{\small $5$} \& 	{\small $6$}  \\
	\end{deptext}
	\depedge{1}{2}{}
	\depedge{4}{3}{}
	\depedge[edge style={red,very thick}]{2}{4}{}
	\depedge[edge style={red,very thick}]{2}{5}{}
	\depedge{5}{6}{}
\end{dependency}} \vspace{-1em}
\setlength{\belowcaptionskip}{-5pt}
\caption{Example of an undirected dependency tree. We observe that the syntactic distance between \word{displeases} and \word{everything} is $2$ (the red path).} 
\label{fig:deptree}
\end{figure}

\paragraph{Syntactic Distance} The syntactic distance between two words in a sentence is, informally, the number of steps between them in an undirected parse tree. Let $\vw = w_1\cdots w_n$ be a sentence of length $n$. 
A parse tree $\vt$ belonging to the sentence $\vw$ is an \emph{undirected} spanning tree of $n$ vertices (with a separate root as a $(n+1)^{\text{th}}$ vertex), each representing a word in the sentence $\vw$. The syntactic distance between two words $w_i$ and $w_j$, denoted $\dist_{\vt}(w_i, w_j)$, is defined as the shortest path from $w_i$ to $w_j$ in the tree $\vt$ where each edge has weight $1$. 
Note that $\dist_\vt(\cdot, \cdot)$ 
is a distance in the technical sense of the word: it is non-negative, symmetric, and satisfies the triangle inequality.

\paragraph{Tree Extraction} 
Converting from syntactic distance to a syntactic tree representation (or vice versa) is trivial and deterministic:
\begin{prop}\label{prop:equiv}
There is a bijection between syntactic distance
and undirected spanning trees.
\end{prop}
\vspace{-.8em}
\begin{proof}
Suppose we have the syntactic distances $\Delta_{\vt}(w_i, w_j)$ for an unknown, undirected spanning tree ${\vt}$. We may uniquely recover that tree by constructing a graph with an edge between $w_i$ and $w_j$ iff $\Delta_{\vt}(w_i, w_j) = 1$.  (This analysis also holds if we have access to only the ordering of the distances between all $|\vw|^2$ pairs of words, rather than the perfect distance calculations---if that were the case, the minimum spanning tree could be computed e.g.\ with \citeauthor{prim}'s.) 
On the other hand, if we have an undirected spanning tree $\vt$ and wish to recover the syntactic distances, we only need to compute the shortest path between each pair of words, with e.g.\ \citeauthor{floyd}--\citeauthor{warshall}, to yield $\Delta_{\vt}(\cdot, \cdot)$ uniquely.
\end{proof}

\newcommand{\nCr}[2]{\,^{#1}C_{#2}}

\section{Probe, Meet Parser}\label{sec:two-parsers}

In this section, we introduce a popular syntactic probe 
and a more traditional parser. 
\looseness=-1

\subsection{The Structural Probe} \label{sec:struct}
\newcite{hewitt} introduce a novel method for approximating the syntactic distance $\dist_\vt(\cdot, \cdot)$ between any two words in a sentence. They christen their method the \defn{structural probe},   
since it is intended to uncover latent syntactic structure in contextual embeddings.\footnote{In actual fact, the structural probe consists of two probes, one used to estimate the syntactic distance between words (which recovers an undirected graph) and another to calculate their depth in the tree (which is used to recover ordering). In this work, we focus exclusively on the former.} 
To do this, they define a parameterised distance function whose parameters are to be learned from data. For a word $w_i$, let $\vh_i \in \R^d$ denote its contextual embedding, where $d$ is the dimensionality of the embeddings from the model we wish to probe, such as $\BERT{}$. \citet{hewitt} define the parameterised
distance function
\begin{align}
    \dB(w_i, w_j) &=  \\ &\sqrt{(\vh_i -\vh_j)^{\top}B^{\top}B\,(\vh_i-\vh_j)}\nonumber
\end{align}
where $B \in \R^{r \times d}$ is to be learned from data, and $r$ is a user-defined hyperparameter. 
The matrix $B^{\top} B$ is positive semi-definite and has rank at most $r$.\footnote{To see this, let $\vx \in \R^d$ be a vector. Then, we have that $\vx^{\top} B^{\top} B \vx = (B \vx)^{\top} (B \vx) = ||B \vx||_2^2 \geq 0$.}

The goal of the structural probe, then, is to find $B$ such that the distance function $\dB(\cdot, \cdot)$ best approximates
$\dist(\cdot, \cdot)$.
If we are to organise our training data into pairs, each consisting of a gold tree $\vt$ and its corresponding sentence $\vw$, we can then define the local loss function as 
\begin{align}\label{eq:hw-local}
    \lhm(B, &\langle \vt, \vw \rangle) = \\
    &\sum_{i=1}^{|\vw|} \sum_{j=i+1}^{|\vw|} \Big| \dist_{\vt}(w_i, w_j) - \dB(w_i, w_j)\Big|  \nonumber
\end{align}
which is then averaged over the entire training
set $\calD = \{\langle \vt^{(k)}, \vw^{(k)}\rangle\}_{k=1}^N$ to create the following global objective
\begin{align}\label{eq:hewitt-global}
    \Lhm(B) &= \sum_{k=1}^N \frac{1}{|\vw^{(k)}|^2} \,\lhm\left(B, \langle \vt^{(k)}, \vw^{(k)} \rangle\right) 
\end{align} 
Dividing the contribution of each local loss by the square of the length of its sentence (the $|\vw^{(k)}|^2$ factor in the denominator) ensures that each sentence makes an equal contribution to the overall objective, to avoid a bias towards the effect of longer sentences. 
This global loss can be minimised computationally using
stochastic gradient descent.\footnote{
\citeauthor{hewitt} found that replacing $\dB(\cdot, \cdot)$ in \cref{eq:hw-local} with ${\dB}(\cdot, \cdot)^2$ yielded better empirical results, so we do the same. For a discussion of this, refer to App.\ A.1 in \citeauthor{hewitt}.  \newcite{coenen2019visualizing} later
offer a theoretical motivation, based on embedding trees in Euclidean space.}

\subsection{A Structured Perceptron Parser}\label{sec:parser}

Given that probe simplicity seemingly refers to parameterisation rather than the design of loss function, 
we infer that swapping the loss function should not be understood as increasing model complexity. With that in mind, here we describe an alternative to the structural probe which learns parameters for the same function $\dB$---a structured perceptron dependency parser, originally introduced in \newcite{mcdonald-etal-2005-non}. 

This parser's loss function works not by predicting every pairwise distance, but instead by predicting the tree based on the current estimation of the distances between each pair of words, then comparing the total weight of that tree to the total weight of the gold tree (based on the current distance predictions). The local perceptron loss is defined as
\begin{align}\label{eq:perceptron}
   \lperc(B, \langle \vt, &\vw \rangle) = 
    \sum_{(i, j) \in \vt} \dB(w_i, w_j) \\ - &\underbrace{\min_{\vt' \in \calT(\vw)} \! \sum_{(i', j') \in \vt'} \dB(w_{i'}, w_{j'})}_{\textit{computed with \citeauthor{prim}'s algorihtm}} \nonumber
\end{align}
When the predicted minimum spanning tree $\vt'$ perfectly matches the gold tree $\vt$, each edge will cancel and this loss will equal zero. Otherwise, it will be positive, since the sum of the predicted distances for the edges in the gold tree will necessarily exceed the sum in the minimum spanning tree. 
The local losses are summed into a global objective:
\begin{equation}\label{eq:global}
    \Lperc(B) = \sum_{k=1}^N \lperc\left(B, \langle \vt^{(k)}, \vw^{(k)} \rangle\right)
\end{equation}
This quantity can also be minimised, again, with a stochastic gradient method.

Though both the structural probe and the structured perceptron parser may seem equivalent under \cref{prop:equiv}, there is a subtle but important difference. 
To minimise the loss in \cref{eq:hw-local}, the structural probe needs to encode (in $\dB$) the rank-ordering of the distances between each pair of words within a sentence. This is not necessarily the case for the structured perceptron. It could minimise the loss in \cref{eq:perceptron} by just encoding each pair of words as ``near'' or ``far''---and \citeauthor{prim}'s algorithm will do the rest.\footnote{One reviewer argued that, by injecting the tree constraint into the model in this manner, we lose the ability to answer the question of whether a probe discovered trees organically. While we believe this is valid, we do not see why the same criticism cannot be levelled against the structural probe---after all, it is trained on the same trees, just processed into pairwise distance matrices. The trees have been obfuscated, to be sure, but they remain in the data.}
\looseness=-1

\section{Experimental Setup}\label{sec:experiments}

\subsection{Processing Results}

\paragraph{Embeddings and Data} 
We experiment on the contextual embeddings in the final hidden layer of the pre-trained multilingual release of $\BERT{}$ \citep{devlin-etal-2019-bert}, and
trained the models on the Universal Dependency %
\citep{nivre-etal-2016-universal} treebands (v2.4). This allows our analysis to be multilingual. 
More specifically, we consider eight typologically diverse languages (Arabic, Basque, Czech, Finnish, Japanese, Korean, Tamil, and Turkish), plus English.
\looseness=-1

\paragraph{Decoding the Predicted Trees}
Having trained a model to find a $\dB(\cdot, \cdot)$ that approximates $\dist_{\vt}(\cdot, \cdot)$, 
it is trivial to decode test sentences into trees (see \cref{prop:equiv}). For an unseen sentence $\vw = w_1\cdots w_n$, 
we compute the $n\times n$ pairwise distance matrix $D$: 
\begin{equation}
    D_{uv} = \begin{cases}
    \dB(w_i, w_v)& \textbf{if } v>u\\
    0              & \textbf{otherwise}
\end{cases}
\end{equation}
We can then compute the predicted tree $\vt$ from $D$ using \citeauthor{prim}'s algorithm, which returns the minimum spanning tree from the predicted distances.

\subsection{Experiments} \label{sec:metrics}
\begin{figure*}[ht]
    \centering
    \begin{minipage}{\columnwidth}
        \centering
        \includegraphics[width=\columnwidth,trim=0.75cm 0.0cm 1.6cm 1.1cm,clip]{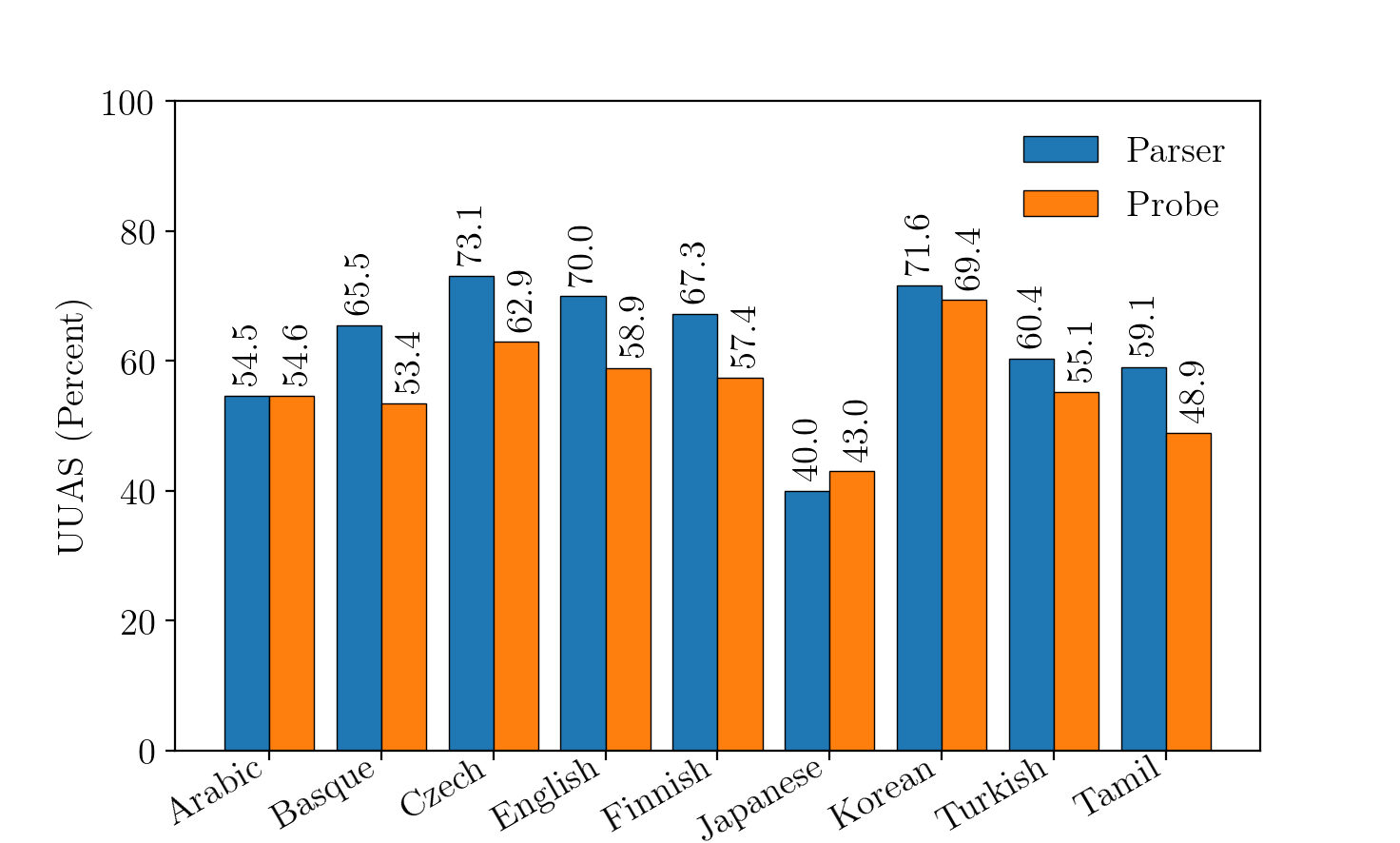}
        \subcaption[first caption.]{UUAS results}\label{fig:uuas}
    \end{minipage}%
    ~~~
    \begin{minipage}{\columnwidth}
        \centering
        \includegraphics[width=\columnwidth,trim=0.8cm 0.0cm 1.6cm 1.1cm,clip]{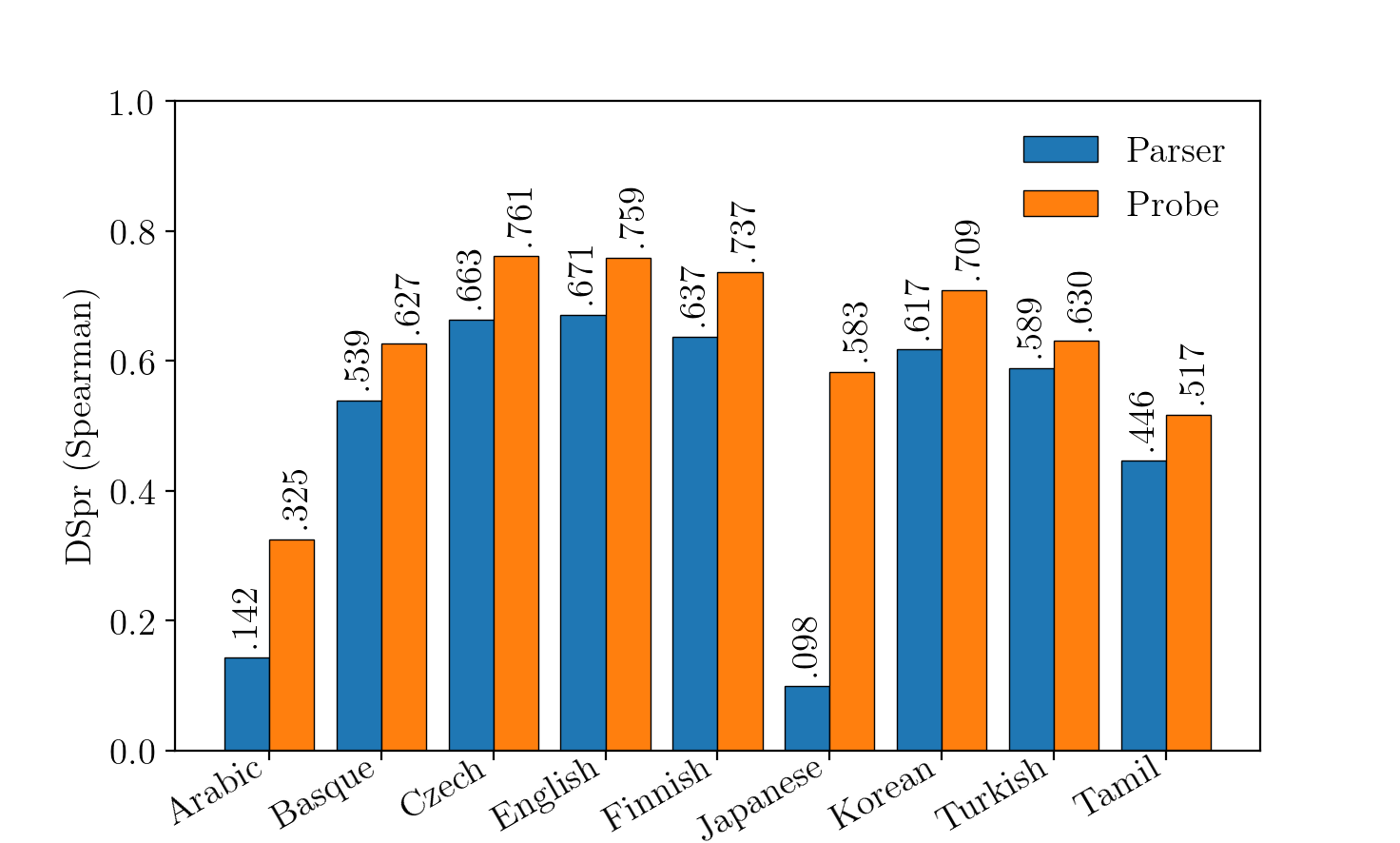}
        \subcaption[first caption.]{DSpr results}\label{fig:DSpr}
    \end{minipage}%
    \setlength{\belowcaptionskip}{-5pt}

    \caption{Results for the metrics in \citet{hewitt}: different metrics, opposite trends.}
    \label{fig:all}
\end{figure*}
To compare the performance of the models, we use both metrics from \citet{hewitt}, plus a new variant of the second. 
\looseness=-1

\paragraph{UUAS} The \textit{undirected unlabeled attachment score} (UUAS) is a standard metric in the parsing literature, which reports the percentage of correctly identified edges in the predicted tree. \looseness=-1

\paragraph{DSpr} The second metric 
is the Spearman rank-order correlation between the predicted distances, which are output from $\dB$, and the gold-standard distances (computable from the gold tree using the \citeauthor{floyd}--\citeauthor{warshall} algorithm). \citeauthor{hewitt} term this metric \defn{distance Spearman} (DSpr). 
While UUAS measures whether the model captures edges in the tree, DSpr considers pairwise distances between \emph{all vertices} in the tree---even those which are not connected in a single step.

\paragraph{DSpr$_{\bm{P+FW}}$}
As a final experiment, we run DSpr again, but first pass each pairwise distance matrix $D$ through \citeauthor{prim}'s 
(to recover the predicted tree) then through \citeauthor{floyd}--\citeauthor{warshall} 
(to recover a new distance matrix, with distances calculated \emph{based on the predicted tree}). 
This post-processing 
would convert a ``near''--``far''  
matrix encoding to a precise rank-order one. This should positively affect the results, in particular for the parser, since that is trained to predict trees which result from the pairwise distance matrix, not the pairwise distance matrix itself.
\looseness=-1

\section{Results}

This section presents results for the structural probe and structured perceptron parser.

\setlength{\belowcaptionskip}{0pt}

\paragraph{UUAS Results}
\cref{fig:uuas} presents UUAS results for both models. 
The parser is the highest performing model on seven of the nine languages. In many of these the difference is substantial---in English, for instance, the parser outperforms the structural probe by $11.1$ UUAS points.\footnote{We used the UD treebanks rather than the Penn-Treebank \citep{marcus-etal-1993-building}, and experimented on the final hidden layer of multilingual $\BERT{}$ using a subset of 12,000 sentences from the larger treebanks. This renders our numbers incomparable to those found in \newcite{hewitt}.}

\paragraph{DSpr Results}
The DSpr results (\cref{fig:DSpr}) 
show the opposite trend: the structural probe outperforms the parser on all languages. The parser performs particularly poorly on Japanese and Arabic,
which is surprising, given that these had the second and third largest sets of training data for $\BERT{}$ respectively (refer to \cref{tab:wikis} in the appendices). We speculate that this may be because in the treebanks used, Japanese and Arabic have a longer average sentence length than other languages.

\setlength{\belowcaptionskip}{-5pt}

\begin{figure}[]
    \centering
    \includegraphics[width=\columnwidth,trim=0.8cm 0.0cm 1.6cm 1.1cm,clip]{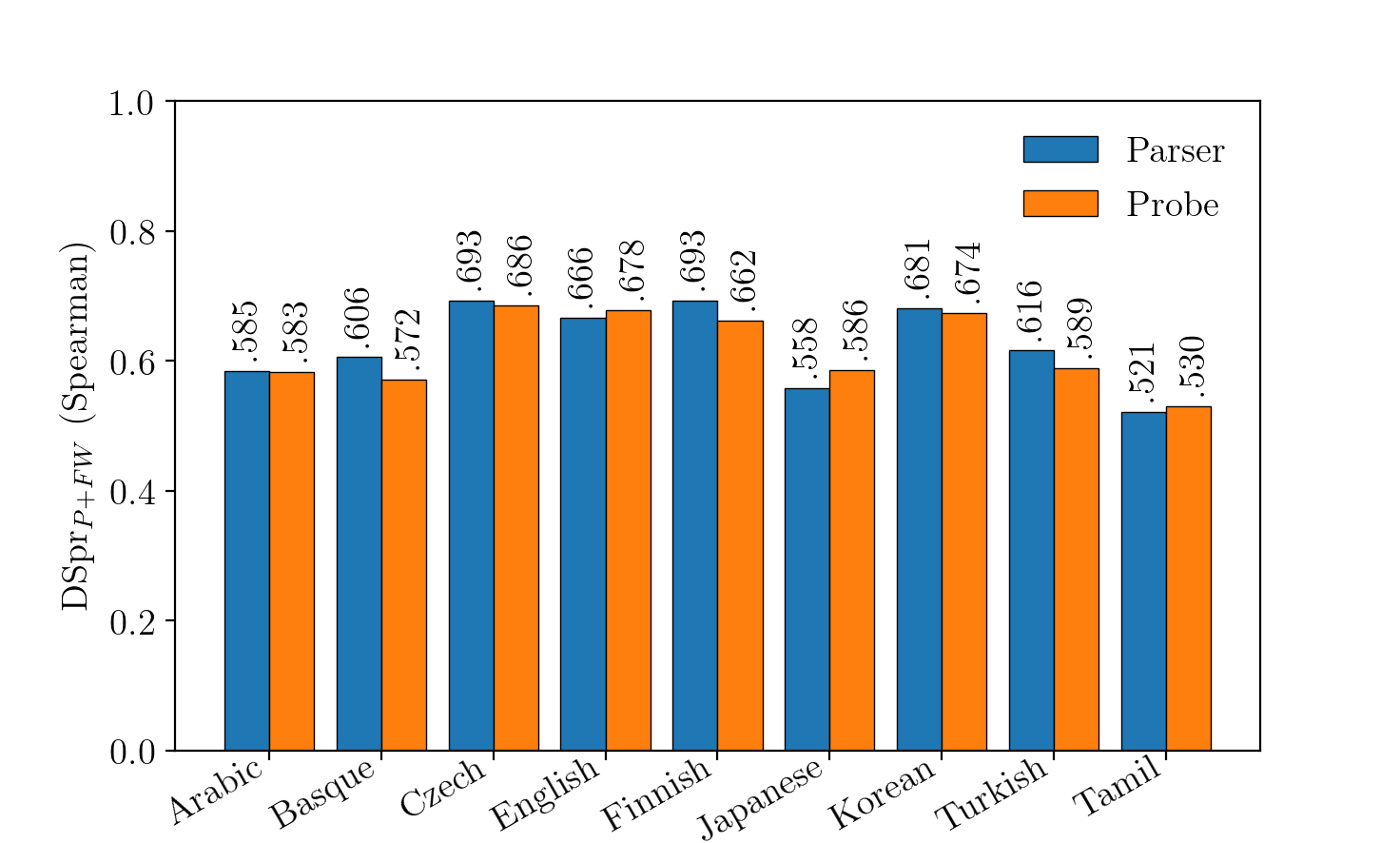}
    \caption{DSpr$_{P+FW}$ results---DSpr following the application of \citeauthor{prim}'s then \citeauthor{floyd}--\citeauthor{warshall} to $D$. 
    }
    \label{fig:DSpr_n}
\end{figure} 

\paragraph{DSpr$_{\bm{P+FW}}$ Results} 
Following the post-processing step, the difference in DSpr (shown in \cref{fig:DSpr_n}) is far less stark than previously suggested---the mean difference between the two across all nine languages is just 0.0006 (in favour of the parser). Notice in particular the improvement for both Arabic and Japanese---where previously (in the vanilla DSpr) the structured perceptron vastly underperformed, the post-processing step closes the gap almost entirely.
Though \Cref{prop:equiv} implies that we do not need to consider the full pairwise output of $\dB$ to account for global properties of tree, this is not totally borne out in our empirical findings, since we do not see the same trend in DSpr$_{P+FW}$ as we do in UUAS. 
If we recover the gold tree, we will have
a perfect correlation with the true syntactic distance---but we do not always recover the gold tree (the UUAS is less than 100\%), and therefore the errors the parser makes are pronounced.

\section{Discussion: Probe v.\ Parser} \label{sec:discussion}

Although we agree that probes should be somehow more constrained in their complexity than models designed to perform well on tasks, we see no reason why being a ``probe'' should necessitate fundamentally different design choices.
It seems clear from our results that how you design a probe has a notable effect on the conclusions one might draw about a representation. 
Our parser was trained to recover trees 
(so it is more attuned to UUAS), 
whilst the structural probe was trained to recover pairwise distances (so it 
is more attuned to DSpr)---viewed this way, our results are not surprising in the least. 

The fundamental question for probe designers, then, is which metric
best captures a linguistic structure believed to be a property of a given representation---in this case, syntactic dependency. 
We suggest that probing research should focus more explicitly on this question---on the development and justification of probing metrics. Once a metric is established and well motivated, a lightweight probe can be developed to determine whether that structure is present in a model. 

If proposing a new metric, however, the burden of proof lies with the researcher to articulate and demonstrate why it is worthwhile. 
Moreoever, this process of exploring which details a new metric is sensitive to (and comparing with existing metrics) ought not be conflated with an analysis of a particular model (e.g.\ $\BERT{}$)---it should be clear whether the results enable us to draw conclusions about a model, or about a means of analysing one.

For syntactic probing, there is certainly no \textit{a-priori} reason why one should prefer 
DSpr to UUAS. If anything, we tentatively recommend UUAS, pending further investigation. The DSpr$_{P+FW}$ 
results show no clear difference between the models, whereas UUAS exhibits a clear trend in favour of the parser, suggesting that 
it may be easier to recover pairwise distances from a good estimate of the tree than vice versa. UUAS also has the advantage that it is well described in the literature (and, in turn, well understood by the research community).

According to UUAS, existing methods were able to identify more syntax in $\BERT{}$ than the structual probe.
In this context, though, we use these results not to give kudos to $\BERT{}$, 
but to argue that the perceptron-based parser is a better tool for syntactic probing.
Excluding differences in parameterisation, the line between what constitutes a probe or a model designed for a particular task is awfully thin, and when it comes to syntactic probing, a  powerful probe seems to look a 
lot like a traditional parser.

\looseness=-1

\section{Conclusion}
We advocate for the position that, beyond some notion of model complexity, there should be no inherent difference between the design of a probe and a model designed for a corresponding task.
We analysed the structural probe \citep{hewitt}, and showed that a simple parser with an identical 
lightweight parameterisation was able to identify more syntax in $\BERT{}$ in seven of nine compared languages under UUAS. 
However, the structural probe outperformed the parser on a novel metric proposed in \newcite{hewitt}, bringing to attention a broader question: how should one choose 
metrics for probing?
In our discussion, we argued that 
if one 
is to propose a new metric,  
they should clearly justify its usage.

\section*{Acknowledgements} 

We thank John Hewitt for engaging wholeheartedly with our work and sharing many helpful insights. 

\bibliography{anthology,acl2020} 
\bibliographystyle{acl_natbib}

\appendix

\section{Training Details}

For all models (separately for each language), we considered three hyperparameters: 
the rank $r$ 
(full rank when $r=768$, since this is the dimensionality of the $\BERT{}$ embeddings), the learning rate, and the dropout rate \cite{JMLR:v15:srivastava14a}. To optimise these, we performed a random search, 
selecting values as judged by loss on the development set. 
When training, we used a batch size of 64 sentences, and employ early stopping after five steps based on loss reduction. 
As the optimiser, we used Adam \cite{DBLP:journals/corr/KingmaB14}. 

For each language, we used the largest available Universal Dependency 2.4 treebank. One-word sentences and sentences of over 50 words were discarded, and the larger treebanks were pruned to 12,000 sentences (in an 8:1:1 data split). 

We use the $\BERT{}$ implementation of \newcite{wolf2019hugging}. Since $\BERT{}$ accepts WordPiece units \citep{wu2016google} rather than words, 
where necessary we averaged the output to get word-level embeddings. This is clearly a na\"{i}ve composition method; improving it would likely strengthen the results for both the probe and the parser.

\section{Multilingual $\bm{\BERT{}}$ Details}\label{sec:bert_details}

 Multilingual $\BERT{}$ has 12 layers, 768 hidden states, and a total of 110M parameters. It was trained on the complete Wikipedia dumps for the 104 languages with the largest Wikipedias. \cref{tab:wikis} reports the size of the Wikipedias for the languages considered in this paper.\footnote{According to \url{https://en.wikipedia.org/wiki/List_of_Wikipedias}, sampled 24/10/19.} Further details of the training can be found on Google Research's GitHub.\footnote{\url{https://github.com/google-research/bert/blob/master/multilingual.md}}

\begin{table}[h]
    \centering
     \begin{tabular}{lr}\toprule
     Language & Articles \\
     \midrule
     Arabic & 1,016,152 \\
     Basque & 342,426 \\
     Czech & 439,467 \\
     English & 5,986,229 \\
     Finnish & 473,729 \\
     Japanese & 1,178,594 \\
     Korean & 476,068 \\
     Tamil & 125,031 \\
     Turkish & 336,380 \\ \bottomrule
     \end{tabular}
    \caption{The number of articles in the Wikipedias of the languages considered.}
    \label{tab:wikis}
\end{table}

\end{document}